\documentclass[runningheads]{llncs}

\usepackage{eccv}

\usepackage{eccvabbrv}
\usepackage{subcaption}
\usepackage{graphicx}
\usepackage{booktabs}
\usepackage{xcolor}
\usepackage{colortbl}
\usepackage{multirow}
\usepackage{caption}  
\usepackage[pagebackref,breaklinks,colorlinks]{hyperref}
\usepackage[accsupp]{axessibility}  

\usepackage{hyperref}
\usepackage{wrapfig}
\usepackage{orcidlink}

\begin{document}

\title{AutoFocus: Uncertainty-Aware Active Visual Search for GUI Grounding} 

\titlerunning{Abbreviated paper title}

\author{Ruilin Yao\inst{1, 2} \and
Shegnwu Xiong\inst{1} \and
Tianyu Zou\inst{1} \and
Shili Xiong\inst{1} \and
Yi Rong\inst{1}}

\authorrunning{F.~Author et al.}

\institute{Wuhan University of Technology \and Institute of Automation, Chinese Academy of Sciences \\
\url{https://github.com/Mr-Bigworth/AutoFocus} \\
\email{{yaoruilin}@whut.edu.cn}}

\maketitle

\begin{abstract}
Vision-Language Models (VLMs) have enabled autonomous GUI agents that translate natural language instructions into executable screen coordinates. However, grounding performance degrades in high-resolution interfaces, where dense layouts and small interactive elements expose a resolution gap between modern displays and model input constraints. Existing zoom-in strategies rely on fixed anchors, heuristic grids, or reinforcement learning, lacking a principled mechanism to adaptively determine where refinement is needed and how much spatial uncertainty should be explored.
We propose AutoFocus, a training-free, uncertainty-aware active visual search framework for GUI grounding. Our key insight is that token-level perplexity in coordinate generation naturally reflects spatial uncertainty. Rather than committing to a single prediction, AutoFocus samples multiple coordinate hypotheses and converts their axial perplexities into an anisotropic gaussian spatial probability field, explicitly modeling directional uncertainty. Based on this field, we generate global and local region proposals and introduce Shape-Aware Zooming to balance tight localization with contextual preservation. A visual prompt-based aggregation step then selects the most consistent prediction via structured comparison.
Extensive experiments on ScreenSpot-Pro and ScreenSpot-V2 demonstrate consistent improvements across both general-purpose and GUI-specialized VLMs. 
\end{abstract}

\section{Introduction}
\label{sec:intro}
The synergy between Vision Language Models (VLMs) and Graphical User Interface (GUI) automation has paved the way for autonomous agents capable of executing complex tasks across digital platforms. These agents \cite{yuan2025segui, huang2025spiritsight, sun2025gui} aim to bridge the ``intention-action gap'' by transforming natural language instructions into precise on-screen operations, such as identifying a specific pixel coordinate to execute an action based on the given textual instruction. However, despite the remarkable reasoning capabilities of frontier VLMs, they still lack precise grounding ability when deployed in professional, high-resolution environments. 

This failure stems from a fundamental resolution mismatch. Modern interfaces operate at $1080p$ or $4K$ resolutions, where critical interactive elements may occupy only a few dozen pixels. In contrast, VLMs typically process resized inputs with constrained visual tokens, limiting their ability to resolve fine-grained details. As a result, a model may semantically identify the correct UI component yet fail to localize it precisely, producing coordinate offsets or missing small icons altogether. This discrepancy exposes a structural bottleneck: spatial grounding requires higher visual fidelity than global semantic reasoning.
The issue is further compounded by the prevailing single-pass grounding paradigm. Most models produce a coordinate prediction in one forward pass, without intrinsic verification or adaptive focus mechanisms. When uncertainty arises, such as in visually crowded layouts, models lack a principled strategy to decide whether refinement is needed, where to re-examine, and how much context to preserve.

\begin{wrapfigure}{t}{0.57\textwidth}
  \centering
  \vspace{-20pt} 
  \includegraphics[width=\linewidth]{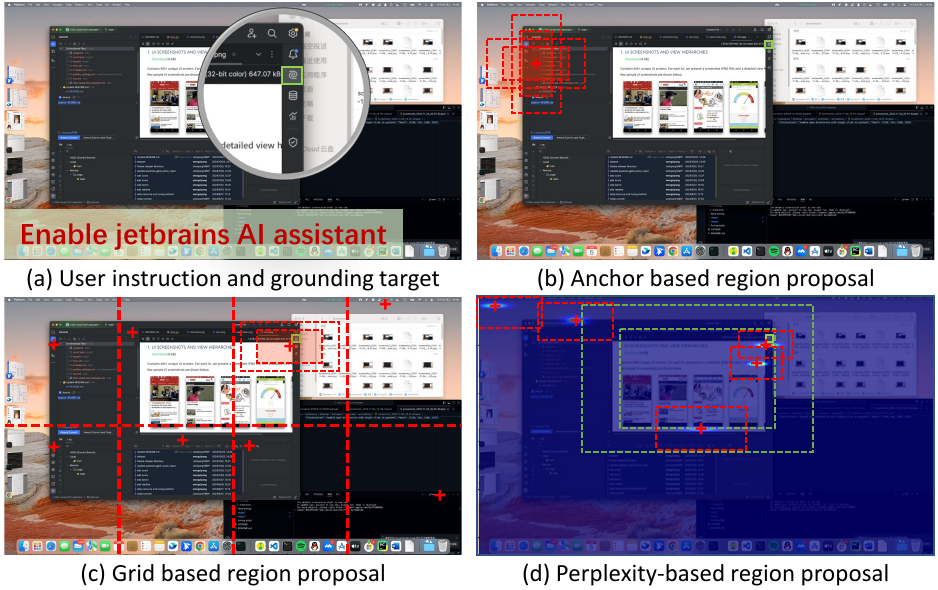}
  \caption{Comparison with other zoom-in methods.}
  \label{fig:intro0}
  \vspace{-18pt} 
\end{wrapfigure}
Recent works have explored multi-stage “zoom-in” strategies to mitigate this problem. However, existing approaches often rely on fixed anchors, heuristic grids, or zooming around a single initial prediction (as shown in \cref{fig:intro0}), limiting adaptability across varying GUI scales \cite{luo2025visual, wu2025dimo, park2025r, jiang2025zoom}. Other methods employ reinforcement learning to optimize cropping policies \cite{lei2025textsc, ye2025gui}, but at the cost of significant training overhead and environmental interaction. These solutions lack a unified and model-intrinsic criterion for adaptive refinement.

\begin{figure}[t]
  \centering
  \includegraphics[width=1\linewidth]{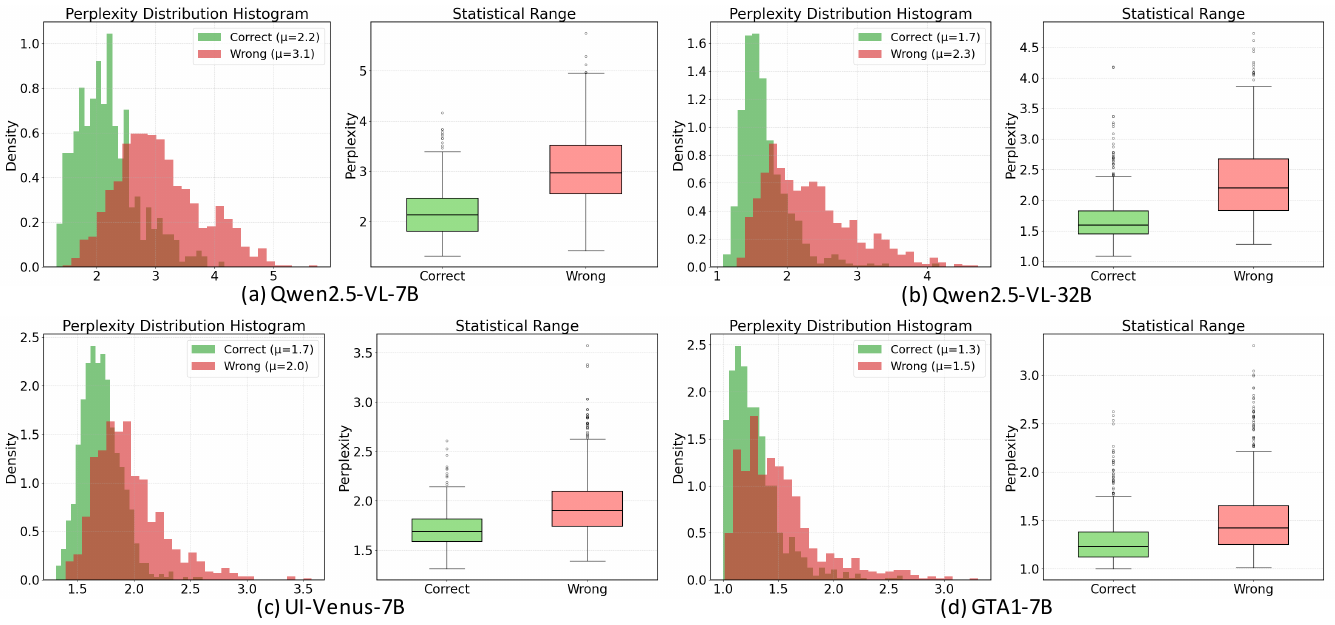}
  \caption{Perplexity (PPL) distribution of grounding results on ScreenSpot-Pro. We compare the PPL across multiple models, with \textcolor{ForestGreen}{green} and \textcolor{red}{red} areas indicating correct and incorrect grounding predictions, respectively. Higher PPL values for incorrect results suggest a strong correlation between model uncertainty and grounding failure.}
  \label{fig:intro1}
  \vspace{-10pt}
\end{figure}

In this work, we propose AutoFocus, a training-free, uncertainty-aware active visual search framework for high-resolution GUI grounding. 
Our key insight is that token-level perplexity (PPL) in coordinate generation correlates with spatial ambiguity. Rather than treating perplexity purely as a language modeling metric, we reinterpret it as an intrinsic uncertainty signal reflecting the model’s confidence in spatial localization. 
We empirically validate this assumption in \cref{fig:intro1}, where we analyze the relationship between token-level perplexity and grounding correctness across multiple backbone models. As shown, incorrect predictions consistently exhibit significantly higher axial perplexity than correct ones, with clear distributional separation in both histogram and boxplot statistics. This trend holds across general-purpose (Qwen2.5-VL \cite{bai2025qwen2}) and GUI-specialized models (UI-Venus \cite{gu2025ui}, GTA1 \cite{yang2025gta1}), indicating that perplexity serves as a reliable intrinsic proxy for spatial uncertainty. 
These findings motivate a principled refinement strategy: convert token-level perplexity into explicit spatial variance and guide adaptive visual search accordingly.

Specifically, AutoFocus first determines whether refinement is necessary based on the model’s initial prediction and its associated uncertainty. If the prediction exhibits high ambiguity, the core Gaussian Dynamic Focusing (GDF) module is activated.
Specifically, we map the $x$- and $y$-axis perplexities into anisotropic variances $(\sigma_x, \sigma_y)$, forming a continuous 2D Gaussian probability field over the high-resolution screenshot. This density field models directional uncertainty, enabling the agent to reason about both where ambiguity lies and how uncertainty spreads spatially. Based on this field, AutoFocus generates global and local region proposals and introduces Shape-Aware Zooming, which adaptively balances tight localization with contextual preservation. Finally, predictions are performed on these region proposals and a structured aggregation stage selects the most consistent prediction.

Unlike rigid anchor-based cropping or training-intensive reinforcement learning, AutoFocus operates entirely at test time and requires no parameter updates. By leveraging intrinsic uncertainty signals already available within the model, AutoFocus dynamically tailors its ``foveal'' search area and provides a principled and computationally controllable refinement mechanism for high-resolution grounding.
Our contributions are summarized as follows:
\begin{itemize}
\item 
We propose AutoFocus, a plug-and-play refinement framework that mitigates the inherent resolution bottleneck in GUI grounding. By conceptualizing grounding as an uncertainty-driven active visual search, AutoFocus empowers VLMs to dynamically re-examine ambiguous interface elements without requiring additional fine-tuning.
\item We introduce Gaussian Dynamic Focusing, which converts token-level perplexity into anisotropic spatial probability fields to construct global-local region proposals, enabling directional and scale-adaptive zooming.
\item We conduct comprehensive evaluations on the ScreenSpot-V2 and ScreenSpot-Pro benchmarks. AutoFocus consistently delivers substantial performance gains across various base models. Notably, our framework reaches 67.8\% accuracy on ScreenSpot-Pro when integrated with UI-Venus-7B.
\end{itemize}

\section{Related Work}

\subsection{GUI Agents}
Recent advancements in Vision Language Models (VLMs) have significantly enhanced GUI automation, enabling agents to navigate graphical user interfaces with increasing sophistication \cite{deng2023mind2web, gou2024navigating, pahuja2025explorer, wang2024mobile, chen2025guicourse, chen2025less}. Prior studies generally adopt text-based reasoning, using structured representations such as HTML or accessibility trees~\cite{koh2024visualwebarena, cao2024spider2}. Some methods also use LLMs to extract structured interface information~\cite{yu2025omniparser} and supplementary textual details for agents input~\cite{zheng2024gpt}. However, as the field moves toward a more ``human-like'' embodiment and the enhancement of the VLMs' visual reasoning capabilities, recent methods emphasize direct visual interpretation of GUI elements from raw pixels~\cite{gou2024navigating, qin2025ui}. 
Despite these strides, they often struggle with high-resolution or ambiguous professional interfaces~\cite{li2025screenspot, wu2024atlas}, where diminutive icons and dense layouts degrade grounding reliability. Furthermore, the prevailing reliance on supervised fine-tuning (SFT) poses limitations in scalability and adaptability, motivating a paradigm shift toward test-time scaling, aims to overcome the need for vast labeled datasets and improve generalization to unseen interfaces. 

\subsection{Visual Test-Time Scaling}
Test-time scaling involves scaling computational resources during inference to enhance model performance~\cite{wang2023self, yao2023tree, ye2024prompt  }. This approach allows AI agents to allocate additional processing power to challenging scenarios, thereby improving perception and reasoning. Inspired by advancements in textual test-time scaling for LLMs (\eg Chain-of-thought prompting \cite{wei2022chain}), recent studies have extended this logic to the visual domain, where reasoning is structured as a sequential trajectory of continuously adjusted visual content \cite{gupta2023visual, zheng2025deepeyes, su2025thinking, wei2026zooming}. 

For instance, Visual Programming \cite{gupta2023visual} utilizes auxiliary spatial cues, such as lines drawn on diagrams, to ground visual reasoning.
Techniques like Vicrop \cite{zhang2025mllms} employ a model's internal attention maps and gradient saliency to automate visual cropping, while DYFO \cite{li2025dyfo} facilitates a bidirectional interaction between LMMs and external ``visual experts'' using Monte Carlo Tree Search (MCTS) to simulate human-centric focus adjustments.
In the specific context of GUI agents, Region-Focus \cite{luo2025visual} introduces an ``image-as-map'' mechanism paired with simplified region selection to improve interactivity. Similarly, Dimo-GUI \cite{wu2025dimo} dynamically refines grounding results by zooming into sub-regions centered on the model's initial—and potentially noisy—coordinate predictions. 
Unlike existing approaches that rely on rigid crops or computationally expensive search algorithms, our framework leverages internal model perplexity as an intrinsic navigation signal, dynamically adapts its search area to transcend the limitations of fixed-region zooming, ensuring high-precision localization even within dense and complex digital interfaces.

\section{Method}

\begin{figure}[tb]
  \centering
  \includegraphics[width=0.99\linewidth]{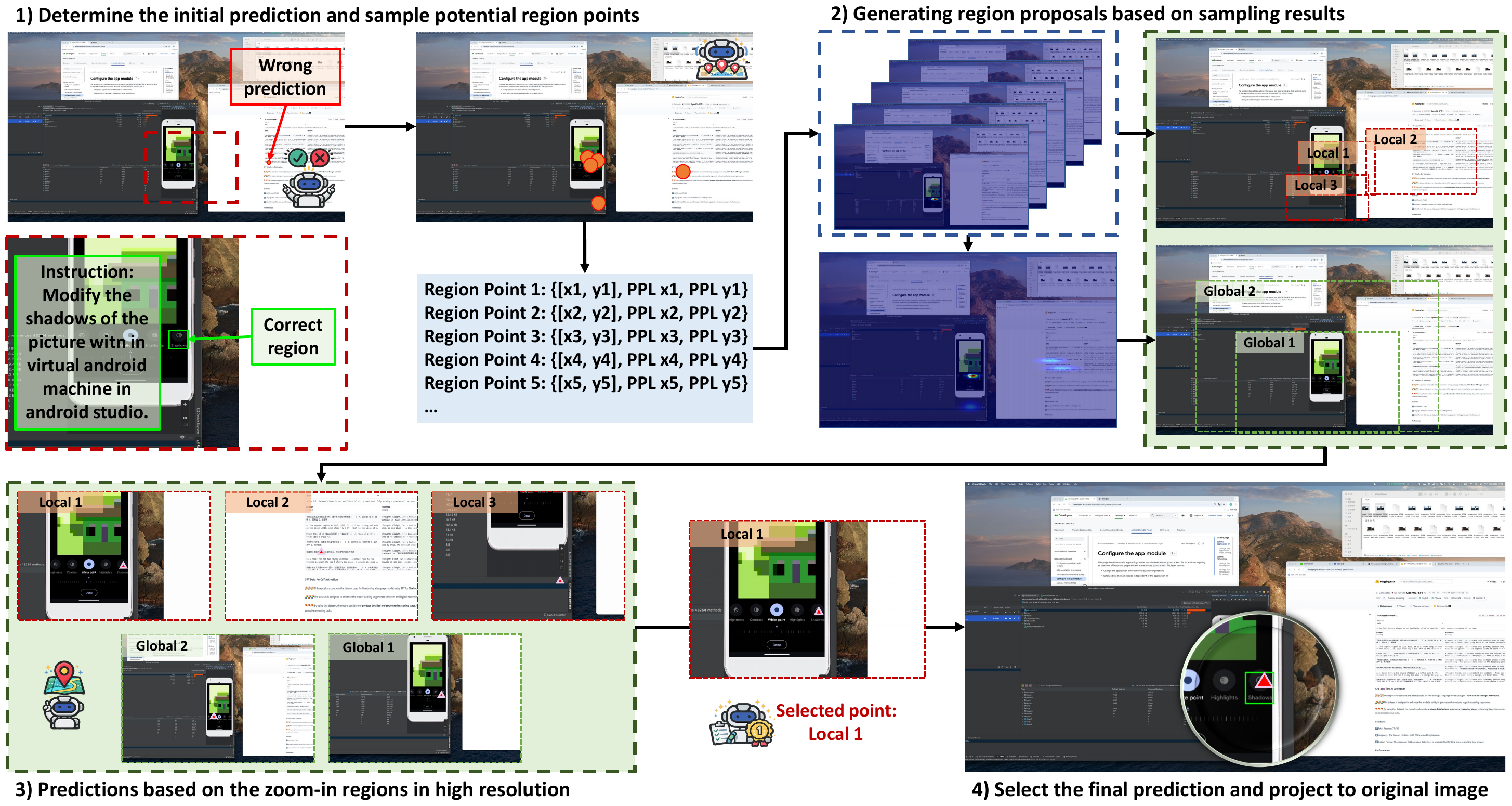}
  \caption{Overview of AutoFocus. The model first samples multiple coordinate hypotheses with axial perplexities from the initial prediction. Gaussian Dynamic Focusing constructs an anisotropic spatial field to generate global and local proposals. High-resolution zoom-in predictions are performed on candidate regions, and a structured aggregation step selects the final coordinate.}
  \label{fig:overall}
\end{figure}

In this section, we present AutoFocus, a training-free, uncertainty-aware active visual search framework designed to bridge the resolution gap in GUI grounding, followed by region proposal generation and visual aggregation. The overall framework is depicted in \cref{fig:overall}.
\subsection{Error-Triggered Active Refinement}

Given a text description and the original image, AutoFocus first performs an initial coarse grounding prediction. Instead of blindly applying zoom-in refinement to every prediction, AutoFocus employs a visual prompt-based self-verification mechanism to determine whether refinement is necessary. This allows the framework to selectively allocate computational resources to uncertain cases while avoiding redundant zoom-in operations for confident predictions.

Specifically, given the initial prediction $p_0 = (x_0, y_0)$, we construct a visual-prompted image $I_{v} = \mathcal{V}(I_{raw}, p_0)$, where $\mathcal{V}$ denotes a drawing operation that overlays a salient visual marker (\eg, a colored point or star \cite{luo2025visual}) at the predicted location. The prompted image is then fed back into the VLM together with a verification query, such as: \textit{``Does the marked point correctly correspond to the target element described in the instruction?''} 

Based on the model’s response, AutoFocus determines whether the prediction is sufficiently reliable. If the model confirms the prediction, the coordinate is directly accepted as the final result. Otherwise, the prediction is considered uncertain, and the Gaussian Dynamic Focusing (GDF) module is activated to construct a spatial uncertainty field and perform targeted zoom-in refinement. This visual prompt-based verification converts the refinement trigger into a structured visual reasoning task, enabling AutoFocus to dynamically and efficiently focus on ambiguous regions.

\subsection{Gaussian Dynamic Focusing}

When visual self-verification determines that the initial prediction is unreliable, AutoFocus activates the Gaussian Dynamic Focusing (GDF) module to perform uncertainty-guided spatial exploration. Unlike conventional grounding methods that rely on a single deterministic prediction, GDF explicitly models the spatial uncertainty inherent in the VLM’s generative process and converts it into a structured spatial probability field. This enables principled region proposal generation that reflects both prediction confidence and directional ambiguity.

GDF consists of three stages: (1) Probabilistic coordinate sampling with axial perplexity estimation, (2) Anisotropic gaussian splatting to construct a continuous spatial density field, and (3) Global-local region proposal generation.
\paragraph{Probabilistic coordinate sampling and axial perplexity.}

Modern VLMs formulate coordinate grounding as an autoregressive token generation process over a discrete spatial vocabulary. Given an image $\mathcal{I}$ and instruction $\mathcal{T}$, the model produces a sequence of coordinate tokens whose likelihood reflects its confidence in the prediction. While greedy decoding yields only a single point estimate, selecting the token with the highest posteriori probability as the final prediction, it discards valuable information about prediction uncertainty.
To recover this information, we perform stochastic decoding with temperature $\tau$ to obtain a set of $N$ sampled coordinates:

\begin{equation}
\mathcal{S} = \{p_i\}_{i=1}^{N}, \quad p_i = (x_i, y_i)
\end{equation}
where each $p_i$ represents the $i$-th sampled coordinate of potential target locations based on the instruction. Each coordinate is generated as a token sequence, which we decompose into independent axial subsequences for the horizontal and vertical dimensions:

\begin{equation}
\mathbf{t}_x = \{t_{x,1}, \dots, t_{x,L_x}\}, \quad
\mathbf{t}_y = \{t_{y,1}, \dots, t_{y,L_y}\}
\end{equation}
where $L_x$ and $L_y$ denote the length of two axial subsequences $\mathbf{t}_x$ and $\mathbf{t}_y$.
Then, we compute axial perplexity as a measure of uncertainty along each spatial dimension:

\begin{equation}
PPL_x^{(i)} =
\exp \left(
-\frac{1}{L_x}
\sum_{j=1}^{L_x}
\log P(t_{x,j} \mid \mathbf{t}_{x,<j}, \mathcal{I}, \mathcal{T})
\right)
\end{equation}

\begin{equation}
PPL_y^{(i)} =
\exp \left(
-\frac{1}{L_y}
\sum_{j=1}^{L_y}
\log P(t_{y,j} \mid \mathbf{t}_{y,<j}, \mathbf{t}_x, \mathcal{I}, \mathcal{T})
\right).
\end{equation}
where $PPL_{x}^{(i)}$ and $PPL_{y}^{(i)}$ denotes the perplexity of the $i$-th sampled coordinate on axis $x$ and $y$. $L_x$ and $L_y$ represent the number of tokens used to encode the $x$ and $y$ coordinates, respectively. 
Perplexity serves as an intrinsic uncertainty signal derived directly from the model’s generative process. Higher perplexity indicates lower confidence and suggests that the true target location may lie within a broader spatial neighborhood.

Importantly, axial decomposition enables directional uncertainty modeling. GUI elements such as text fields and sliders often exhibit anisotropic geometry, where uncertainty differs across axes. Modeling these independently allows more precise spatial reasoning.
\paragraph{Anisotropic gaussian splatting.}
To bridge the gap between discrete coordinate samples and the underlying continuous spatial distribution, we project the sampled hypotheses into a continuous probability field via Anisotropic Gaussian Splatting. Each sampled coordinate is modeled as an anisotropic Gaussian kernel:
\begin{equation}
\mathcal{N}_i(p) =
\exp \left(
-\frac{1}{2}
(p - \mu_i)^T \Sigma_i^{-1} (p - \mu_i)
\right),
\end{equation}
where $\mu_i = (x_i, y_i)$ represents the sampled centroid. The covariance matrix $\Sigma_i$ is decoupled to explicitly reflect the directional uncertainty of the model.
\begin{equation}
\mu_i = (x_i, y_i),
\quad
\Sigma_i =
\begin{bmatrix}
\sigma_{x,i}^2 & 0 \\
0 & \sigma_{y,i}^2
\end{bmatrix},
\end{equation}
and the covariance is derived from axial perplexity:
\begin{equation}
\sigma_{z,i} =
\beta
\cdot PPL_z^{(i)},
\quad z \in \{x,y\}.
\end{equation}
Here, the hyperparameter $\beta$ modulates the exploration scale, allowing the kernel to expand along axes where the model exhibits higher perplexity. This anisotropic formulation is critical for capturing the structural priors of GUI elements, such as elongated text fields or narrow navigation bars.Finally, the individual kernels are synthesized into a global spatial density field $M(p)$ through a weighted aggregation:
\begin{equation}
M(p) = \sum_{i=1}^{N} w_i , \mathcal{N}_i(p).
\end{equation}The influence of each sample is governed by its relative confidence $w_i$, computed as the softmax-normalized negative sentence-level perplexity of the $i$-th prediction:
\begin{equation}
w_i = \frac{\exp(-PPL_i)}{\sum^N_j \exp(-PPL_j)}.
\end{equation}
By prioritizing semantically stable samples while accounting for spatial uncertainty, the resulting density field $M(p)$ provides a robust foundation for identifying the most salient candidate regions for zoom-in refinement.
\paragraph{Global-local region proposal generation.}
Based on the density field, we generate region proposals at two complementary scales.

\textbf{Local proposals.}
Each Gaussian kernel defines a spatial hypothesis $B_i$ representing the $3\sigma$ confidence interval of the prediction:
\begin{equation}
B_i =
[x_i - 3\sigma_{x,i}, \; x_i + 3\sigma_{x,i}]
\times
[y_i - 3\sigma_{y,i}, \; y_i + 3\sigma_{y,i}].
\end{equation}
We then apply Non-Maximum Suppression (NMS) to prune redundant proposals, yielding the $K_{local}$ most salient regions. The confidence score for NMS is derived from the model's total response perplexity, ensuring the refinement stage focuses on the most semantically stable predictions.

\textbf{Global proposals.}
The global region proposal is derived from the aggregated spatial density field $M(p)$, which encodes the overall uncertainty-aware distribution over the image plane. We approximate this density field using moment matching and compute the mean and covariance as:

\begin{equation}
\mu =
\frac{\sum_{p} M(p)\, p}{\sum_{p} M(p)},
\quad
\Sigma =
\frac{\sum_{p} M(p)\,(p-\mu)(p-\mu)^T}{\sum_{p} M(p)}
\end{equation}

The center of the global proposal is defined as the mean $\mu$, while its width and height are determined by the standard deviations along each axis:

\begin{equation}
W = \alpha \sigma_x, \quad H = \alpha \sigma_y
\end{equation}
where $\sigma_x = \sqrt{\Sigma_{xx}}$ and $\sigma_y = \sqrt{\Sigma_{yy}}$, and $\alpha$ is a scale factor controlling the coverage of the global proposal region. When multiple global proposals are required, different scale factors $k$ can be used to construct a hierarchy of $K_{global}$ regions capturing varying levels of spatial context.

Since the resulting proposal may extend beyond image boundaries, we apply a boundary-aware adjustment that translates the proposal toward the image center while preserving its dimensions. This ensures that the proposal remains fully contained within the valid image region without altering its uncertainty-derived spatial extent.
To further balance precision and contextual coverage, we introduce Shape-Aware Zooming (SAZ), which interpolates between anisotropic and square crops:

\begin{equation}
W_{final}
=
W_{raw}
+
\lambda
(\max(W_{raw},H_{raw}) - W_{raw}),
\end{equation}
\begin{equation}
H_{final}
=
H_{raw}
+
\lambda
(\max(W_{raw},H_{raw}) - H_{raw}),
\end{equation}
where $\lambda \in [0,1]$ controls the trade-off between tight localization and contextual completeness.

Together, these mechanisms enable AutoFocus to perform uncertainty-aware, geometry-adaptive zoom-in refinement, significantly improving grounding accuracy in high-resolution GUI environments.

\subsection{Action Prediction and Aggregation}
Given the set of global and local region proposals generated by GDF, AutoFocus performs a second-pass high-resolution grounding step followed by a visual aggregation stage to produce the final action coordinate.

For each proposal region $R_j$, where $j \in \{1, \dots, K\}$ and $K = K_{\text{global}} + K_{\text{local}}$, we independently apply the VLM to perform refined grounding under a zoomed-in view. Specifically, each region $R_j$ is cropped from the original image and resized to the model’s standard input resolution, enabling the VLM to access fine-grained visual details that may be indistinguishable at the original scale. The resized crop, together with the original instruction $\mathcal{T}$, is fed into the VLM to produce a refined coordinate prediction:
$\hat{c}_j = (\hat{x}_j, \hat{y}_j),$
where $(\hat{x}_j, \hat{y}_j)$ denotes the predicted coordinate in the local coordinate system of the resized crop.

To ensure consistency across different proposals, each refined prediction is projected back into the original image coordinate system. Let $(x_j^{\min}, y_j^{\min})$ denote the top-left corner of region $R_j$ in the original image, and let $(s_{x,j}, s_{y,j})$ denote the scaling factors between the original crop and the resized input along the horizontal and vertical axes, respectively.
The global coordinate is computed as:
\begin{equation}
c_j =
\left(
x_j^{min} + \frac{\hat{x}_j}{s_{x,j}},\;
y_j^{min} + \frac{\hat{y}_j}{s_{y,j}}
\right).
\end{equation}
This yields a set of candidate coordinates $\{c_j\}_{j=1}^{K}$, each corresponding to a refined hypothesis derived from a distinct uncertainty-guided region.

Instead of selecting the final coordinate based solely on numerical confidence, AutoFocus formulates the aggregation step as a multi-image visual reasoning task. For each candidate coordinate $c_j$, we construct an annotated image $\tilde{\mathcal{I}}_j = \mathcal{V}(\mathcal{I}, c_j)$ by overlaying a visual marker at the corresponding location on the original image. This results in a set of annotated images:
\begin{equation}
\tilde{\mathcal{I}} = \{\tilde{\mathcal{I}}_1, \tilde{\mathcal{I}}_2, \dots, \tilde{\mathcal{I}}_K\}.
\end{equation}
These annotated images are jointly provided to the VLM together with the original instruction $\mathcal{T}$, leveraging the model’s native multi-image reasoning capability. The model is prompted to select the image whose marked location best corresponds to the target element. Formally, the VLM produces a categorical distribution over the candidate images:
\begin{equation}
j^* =
\arg\max_j
P(j \mid \tilde{\mathcal{I}}_1,\dots,\tilde{\mathcal{I}}_K,\mathcal{T}),
\end{equation}
and the final predicted coordinate is defined as:
\begin{equation}
c^* = c_{j^*},
\end{equation}
where $j^*$ corresponds to the candidate whose marked position best aligns with the instruction. 
This aggregation strategy transforms the final grounding step into a structured comparison task, allowing the model to leverage contextual reasoning across multiple hypotheses. By converting numerical coordinate predictions into explicit visual prompts, the model can more reliably resolve ambiguities and select the most semantically consistent target location.

\section{Experiments}
\label{sec:exp}
\subsubsection{Base Models.} In the following experiments, we integrate our proposed AutoFocus pipeline with four representative Vision-Language Models (VLMs): 
(i) \textbf{UI-TARS} \cite{qin2025ui}, a model specifically trained for GUI grounding and interaction; 
(ii) \textbf{Qwen2.5-VL} \cite{bai2025qwen2}, a strong general-purpose multimodal foundation model; and 
(iii) \textbf{GTA1-Qwen} \cite{yang2025gta1} and \textbf{UI-Venus} \cite{gu2025ui}, two GUI-specialized agent built upon Qwen2.5-VL with reinforcement learning tailored for UI interaction tasks. These models span both open-domain visual reasoning and specialized GUI grounding paradigms, enabling us to assess whether AutoFocus generalizes beyond task-specialized models.
\subsubsection{Hyperparameter.}
We set the sampling temperature to $\tau = 0.75$ and top-p=1. For region proposal generation, we retain $K_{\text{local}} = 3$ local proposals and $K_{\text{global}} = 2$ global proposals.
These hyperparameters, including $\beta$ and $\alpha$, are empirically determined based on the model's intrinsic perplexity (PPL) distribution and the layout density of different datasets.
For most models, including Qwen-series, UI-TARS and UI-Venus, we use $\beta = 50$. For GTA1-series models, which exhibit systematically lower perplexity and sharper confidence distributions, we use a larger value $\beta = 80$ to maintain a comparable exploration range.
For global proposal construction, we use scale factors $\alpha \in \{5, 8\}$ to capture uncertainty at multiple spatial extents. We use pink stars as visual markers. The squareness factor is set to $\lambda = 0.5$ to balance spatial precision and contextual coverage. Finally, all region proposals are constrained to have a minimum crop size of $336 \times 336$ pixels to ensure sufficient visual detail for reliable refinement.

\subsubsection{Benchmarks.} We evaluate our method on GUI grounding benchmarks covering both OS-level operation tasks and Web interaction scenarios. Specifically, we adopt \textbf{ScreenSpot-Pro}~\cite{li2025screenspot} and \textbf{ScreenSpot-V2} \cite{wu2024atlas}, and report performance using the standard \emph{Accuracy} metric.

\subsubsection{Mitigating self-consistency bias.} AutoFocus is model-agnostic and explicitly supports decoupling the predictor, verifier, and aggregator. In our experiments, we instantiate both the error-triggered verification and the aggregation module using a Qwen2.5-VL of the similar model size as the base model to keep the evaluation controlled and avoid introducing extra capacity. 
\subsection{Main Results}
\begin{table}[!ht]
	\setlength{\tabcolsep}{3.5pt} %
	\caption{Comparison of various models on ScreenSpot-Pro~\cite{li2025screenspot}. The final average scores are in \textbf{bold}. Results of the three base models are highlighted in \colorbox{blue!15}{blue}, and their corresponding AutoFocus-enhanced variants are highlighted in \colorbox{red!15}{red}.}
	\label{tab:screenspot_pro_results}
	\centering
	\resizebox{\linewidth}{!}{
		\begin{tabular}{l|ccc|ccc|ccc|ccc|ccc|ccc|ccc}
			\toprule
			\textbf{Method} & \multicolumn{3}{c|}{\textbf{Development}} & \multicolumn{3}{c|}{\textbf{Creative}} & \multicolumn{3}{c|}{\textbf{CAD}} & \multicolumn{3}{c|}{\textbf{Scientific}} & \multicolumn{3}{c|}{\textbf{Office}} & \multicolumn{3}{c|}{\textbf{OS}} & \multicolumn{3}{c}{\textbf{Avg}} \\
			\cmidrule(lr){2-4} \cmidrule(lr){5-7} \cmidrule(lr){8-10} \cmidrule(lr){11-13} \cmidrule(lr){14-16} \cmidrule(lr){17-19} \cmidrule(lr){20-22}
			& \textbf{text} & \textbf{icon} & \textbf{avg} & \textbf{text} & \textbf{icon} & \textbf{avg} & \textbf{text} & \textbf{icon} & \textbf{avg} & \textbf{text} & \textbf{icon} & \textbf{avg} & \textbf{text} & \textbf{icon} & \textbf{avg} & \textbf{text} & \textbf{icon} & \textbf{avg} & \textbf{text} & \textbf{icon} & \textbf{avg} \\
			\midrule
			GPT-4o \cite{hurst2024gpt} & 1.3 & 0.0 & 0.7 & 1.0 & 0.0 & 0.6 & 2.0 & 0.0 & 1.5 & 2.1 & 0.0 & 1.2 & 1.1 & 0.0 & 0.9 & 0.0 & 0.0 & 0.0 & 1.3 & 0.0 & \textbf{0.8} \\
			Seed-1.5-VL \cite{guo2025seed1}  & -   & -   & 53.8	 & -   & -   & 59.2	  & -   & -   & 59.0 & -   & -   &	61.4  & -   & -   &	74.8  & -   & -   &	60.2 & -   & -   &	\textbf{60.9} \\
			UI-TARS-1.5~\cite{qin2025ui} & -   & -   & 63.9 & -   & -   & 50.4	& -   & -   & 58.2	& -   & -   & 69.3 & -   & -   &	79.6 & -   & -   &	51.0 & -   & -   &	\textbf{61.6} \\
			\midrule
			SeeClick \cite{cheng2024seeclick} & 0.6 & 0.0 & 0.3 & 1.0 & 0.0 & 0.6 & 2.5 & 0.0 & 1.9 & 3.5 & 0.0 & 2.0 & 1.1 & 0.0 & 0.5 & 2.8 & 0.0 & 1.5 & 1.8 & 0.0 & \textbf{1.1} \\
			Qwen2-VL-7B \cite{wang2024qwen2} & 2.6 & 0.0 & 1.3 & 1.5 & 0.0 & 0.9 & 0.5 & 0.0 & 0.4 & 6.3 & 0.0 & 3.5 & 3.4 & 1.9 & 3.0 & 0.9 & 0.0 & 0.5 & 2.5 & 0.2 & \textbf{1.6} \\
			OS-Atlas-4B \cite{wu2024atlas} & 7.1 & 0.0 & 3.7 & 3.0 & 1.4 & 2.3 & 2.0 & 0.0 & 1.5 & 9.0 & 5.5 & 7.5 & 5.1 & 3.8 & 4.4 & 5.6 & 0.0 & 3.1 & 5.0 & 1.7 & \textbf{3.7} \\
			ShowUI-2B \cite{lin2025showui} & 16.9 & 1.4 & 9.4 & 9.1 & 0.0 & 5.3 & 2.5 & 0.0 & 1.9 & 13.2 & 7.3 & 10.6 & 15.3 & 7.5 & 13.5 & 10.3 & 2.2 & 6.6 & 10.8 & 2.6 & \textbf{7.7} \\
			CogAgent-18B \cite{hong2024cogagent}& 14.9 & 0.7 & 8.0 & 9.6 & 0.0 & 5.6 & 7.1 & 3.1 & 6.1 & 22.2 & 1.8 & 13.4 & 13.0 & 0.0 & 6.5 & 5.6 & 0.0 & 3.1 & 12.0 & 0.8 & \textbf{7.7} \\
			Aria-UI \cite{yang2025aria}& 16.2 & 0.0 & 8.4 & 23.7 & 2.1 & 14.7 & 7.6 & 1.6 & 6.1 & 27.1 & 6.4 & 18.1 & 20.3 & 1.9 & 16.1 & 4.7 & 0.0 & 2.6 & 17.1 & 2.0 & \textbf{11.3} \\
			OS-Atlas-7B \cite{wu2024atlas}  & 33.1 & 1.4 & 17.7 & 28.8 & 2.8 & 17.9 & 12.2 & 4.7 & 10.3 & 37.5 & 7.3 & 24.4 & 33.9 & 5.7 & 27.4 & 27.1 & 4.5 & 16.8 & 28.1 & 4.0 & \textbf{18.9} \\
			InfiGUI-R1-3B \cite{liu2025infigui} & 51.3 & 12.4 & 32.4 & 44.9 & 7.0 & 29.0 & 33.0 & 14.1 & 28.4 & 58.3 & 20.0 & 41.7 & 65.5 & 28.3 & 57.0 & 43.9 & 12.4 & 29.6 & 49.1 & 14.1 & \textbf{35.7} \\
			InfiGUI-G1-3B \cite{liu2025infigui-g1} & 64.9 & 20.0 & - & 51.5 & 16.8 & - &50.8 & 25.0 & - & 68.8 & 32.7 & - & 70.6 & 32.1 & - & 49.5 & 15.7 & -& -&-& \textbf{45.2}\\
			SE-GUI-3B~\cite{yuan2025enhancing} & 55.8 & 7.6 & 35.1 & 47.0 & 4.9 & 29.0 & 38.1 & 12.5 & 31.8 & 61.8 & 16.4 & 43.3 & 59.9 & 24.5 & 50.9 & 40.2 &  12.4 & 25.5 & 50.4 & 11.8 & \textbf{35.9} \\
			Jedi-3B \cite{xie2025scaling} & 61.0 & 13.8 & 38.1 & 53.5 & 8.4 & 34.6 & 27.4 & 9.4 & 23.0 & 54.2 & 18.2 & 38.6 & 64.4 & 32.1 & 57.0 & 38.3 & 9.0 & 25.0 & 49.8 & 13.7 & \textbf{36.1} \\
			GUI-G1-3B \cite{zhou2025gui}  & 50.7 & 10.3 & 31.1 & 36.6 & 11.9 & 26.6 & 39.6 & 9.4  & 32.2 & 61.8 & 30.0 & 48.0 & 67.2 & 32.1 & 59.1 & 23.5 & 10.6 & 16.1 & 49.5 & 16.8 & \textbf{37.1} \\
			RegionFocus-7B \cite{luo2025visual} & 53.2 & 3.4 & 29.1 & 42.9 & 4.9 & 27.0 & 28.4 & 3.1 & 22.2 & 56.9 & 10.9 & 37.0 & 59.9 & 24.5 & 51.7 & 41.1 & 15.7 & 29.6 & 46.6 & 8.8 & \textbf{32.1}\\
			RegionFocus-72B \cite{luo2025visual} & 75.3 & 25.5 & 51.2 & 76.3 & 30.8 & 57.2 & 71.6 & 28.1 & 60.9 & 87.5 & 39.1 & 66.5 & 87.0 & 60.4 & 80.9 & 74.8 & 36.0 & 57.1 & 78.6 & 34.1 & \textbf{61.6}\\
			\midrule
			\rowcolor{blue!15} UI-TARS-7B \cite{qin2025ui}& 58.4 & 12.4 & 36.1 & 50.0 & 9.1 & 32.8 & 20.8 & 9.4 & 18.0 & 63.9 & 31.8 & 50.0 & 63.3 & 20.8 & 53.5 & 30.8 & 16.9 & 24.5 & 47.8 & 16.2 & \textbf{35.7}\\
			\rowcolor{red!15} + \emph{AutoFocus} & 64.9 & 13.1 & 39.4 & 63.6 & 14.7 & 43.1 & 43.7 & 14.1 & 37.2 & 78.7 & 30.3 & 57.6 & 75.7 & 35.8 & 66.5 & 57.7 & 20.8 & 40.3 & 63.2 & 18.1 & \textbf{45.1}\\
			\rowcolor{blue!15} UI-TARS-72B \cite{qin2025ui}& 63.0 & 17.3 & 40.8 & 57.1 & 15.4 & 39.6 & 18.8 & 12.5 & 17.2 & 64.6 & 20.9 & 45.7 & 63.3 & 26.4 & 54.8 & 42.1 & 15.7 & 30.1 & 50.9 & 17.5 & \textbf{38.1} \\
			\rowcolor{red!15} + \emph{AutoFocus} & 74.5 & 28.7 & 51.9 & 65.1 & 23.9 & 47.1 & 45.7 & 14.1 & 38.5 & 78.7 & 30.3 & 57.6 & 83.4 & 37.8 & 70.5 & 60.7 & 19.1 & 41.8 & 66.0 & 24.6 & \textbf{52.6}\\
			\midrule
			\rowcolor{blue!15} Qwen2.5-VL-7B \cite{bai2025qwen2} & 45.5 & 1.4 & 24.1 & 32.8 & 6.3 & 21.7 & 22.3 & 6.2 & 18.4 & 50.7 & 7.3 & 31.9 & 52.5 & 15.1 & 43.9 & 36.4 & 10.1 & 24.5 & 39.3 & 6.6 & \textbf{26.8}\\
			\rowcolor{red!15} + \emph{AutoFocus} & 57.5 & 6.4 & 33.1 & 46.8 & 10.0 & 31.1 & 36.2 & 7.9 & 29.3 & 58.3 & 11.0 & 37.9 & 62.3 & 23.5 & 53.5 & 40.2 & 16.9 & 29.6 & 50.2 & 11.3 & \textbf{35.4}  \\
			\rowcolor{blue!15} Qwen2.5-VL-72B \cite{bai2025qwen2} & 66.2 & 13.8 & 40.8 & 64.6 & 15.4 & 44.0 & 47.7 & 12.5 & 39.1 & 78.5 & 29.1 & 57.1 & 74.6 & 37.7 & 66.1 & 60.7 & 22.5 & 43.4 & 64.9 & 20.2 & \textbf{47.8}\\
			\rowcolor{red!15} + \emph{AutoFocus} & 81.2 & 38.1 & 61.1 & 77.3 & 36.5 & 60.6 & 73.1 & 29.0 & 62.4 & 91.0 & 39.3 & 68.9 & 88.6 & 62.7 & 82.8 & 75.2 & 34.5 & 56.8 & 80.9 & 38.6 & \textbf{65.1} \\
			\midrule
			\rowcolor{blue!15} GTA1-Qwen-7B \cite{yang2025gta1}& 66.9   & 20.7   & 44.5	& 62.6   & 18.2   & 44.0	& 53.3   & 17.2   & 44.4   & 76.4 & 31.8   &	57.1 & 82.5   & 50.9   &	75.2 & 48.6  & 25.9   &	38.3 & 65.5   & 25.2   &	\textbf{50.1} \\
			\rowcolor{red!15} + \emph{AutoFocus} & 70.5 & 23.8 & 47.3 & 65.2 & 18.2 & 45.5 & 58.9 & 26.6 & 51.0 & 78.5 & 30.0 & 57.5 & 84.1 & 52.6 & 76.4 & 55.6 & 30.7 & 44.3 & 69.5 & 28.9 & \textbf{54.8}  \\
			\rowcolor{blue!15} GTA1-Qwen-32B \cite{yang2025gta1} & 82.5& 28.3& 56.2& 69.2& 14.7& 46.3& 43.7& 23.4& 38.7& 79.9& 31.8& 59.1& 80.8& 43.4& 72.2& 70.1& 32.6& 53.1& 69.9& 27.2&\textbf{53.6}\\
			\rowcolor{red!15} + \emph{AutoFocus} &
			87.0 & 33.8 & 61.2 & 73.2 & 25.2 & 53.1 & 59.9 & 31.2 & 52.9 & 84.7 & 34.5 & 63.0 & 85.9 & 49.1 & 77.4 & 75.7 & 51.4 & 65.2 & 77.0 & 37.3 & \textbf{63.0}\\
            \midrule
            \rowcolor{blue!15} UI-Venus-7B \cite{gu2025ui} &  73.7 &24.1& 49.8 &61.6 &14.7 &41.9 &58.9 &18.8 &49.0 &76.4 &33.6 &57.9 &75.1 &43.4& 67.8& 49.5 &22.5& 37.2& 66.3& 24.5& \textbf{50.3}\\
            \rowcolor{red!15} + \emph{AutoFocus} &
			84.7 & 35.7 & 63.6 & 75.5 & 37.0 & 60.2 & 82.2 & 43.8 & 73.0 & 90.5 & 51.2 & 75.8 & 88.6 & 59.2 & 83.3 & 78.1 & 39.7 & 61.5 & 82.1 & 41.9 & \textbf{67.8}\\
			\bottomrule
	\end{tabular}}
\end{table}
\subsubsection{ScreenSpot-Pro (SS-Pro).}
\cref{tab:screenspot_pro_results} reports results on the challenging ScreenSpot-Pro benchmark, which features high-resolution professional software interfaces with dense layouts and small interactive elements. AutoFocus consistently improves all base models across every domain category (Development, Creative, CAD, Scientific, Office, and OS). As detailed in the results, AutoFocus enhances UI-TARS-7B by 9.4\% absolute (from 35.7\% to 45.1\%), a scaling effect that becomes even more pronounced in the 72B variant, where accuracy surges from 38.1\% to 52.6\% (+14.5\%). Similar gains are observed for Qwen2.5-VL family: the 7B variant rises from 26.8\% to 35.4\%, while the 72B model improves substantially from 47.8\% to 65.1\% (+17.3). 

Notably, Qwen2.5-VL-72B + AutoFocus achieves an average accuracy of 65.1\%, surpassing strong zoom-based baselines such as RegionFocus-72B (61.6\%). The framework also proves highly effective for RL-tuned agents, pushing GTA1-Qwen-32B to 63.0\% accuracy (+9.4\%). Furthermore, the GUI-specialized UI-Venus-7B experiences a dramatic performance leap, with its average accuracy surging from 50.3\% to 67.8\% (+17.5), demonstrating that even models already optimized for UI grounding benefit profoundly from our uncertainty-guided visual refinement strategy. Improvements are particularly pronounced in icon grounding and small-object-intensive domains (\eg, CAD and Office), demonstrating that uncertainty-aware zooming effectively mitigates resolution-induced localization errors. These results confirm that AutoFocus is model-agnostic and significantly enhances both general-purpose and GUI-specialized VLMs without additional training.

\subsubsection{ScreenSpot-V2 (SS-V2).}
\cref{tab:screenspot_v2_comparison} presents results on ScreenSpot-V2 across mobile, desktop, and web environments. Qwen2.5-VL-7B + AutoFocus achieves an overall accuracy of 93.7\%, competitive with strong 72B baselines. Qwen2.5-VL-72B + AutoFocus further reaches 95.5\%, outperforming the baseline Qwen2.5-VL-72B (94.0\%). Notably, even with a base model that already achieves high-saturation accuracy, our framework provides a substantial boost, raising the overall average from 94.1\% to 95.3\%. 
Gains are consistent across both Text and Icon/Widget categories, with particularly strong improvements in desktop and web icon grounding. Together, these results demonstrate that AutoFocus delivers robust and scalable improvements across both high-resolution professional interfaces and general GUI environments, validating the effectiveness of uncertainty-guided active visual refinement.

\begin{table}[!t]
    \caption{Comparison with state-of-the-art methods on ScreenSpot-V2 \cite{wu2024atlas} across mobile, desktop, and web domains. We report grounding accuracy (\%) categorized by grounding target type: Text, Icon/Widget, and the overall Average (Avg). The final average scores are in \textbf{bold}. The base model results are highlighted in \colorbox{blue!15}{blue}, and their corresponding AutoFocus-enhanced variants are highlighted in \colorbox{red!15}{red}.}
    \centering
    \resizebox{0.8\linewidth}{!}{
    \small
    \begin{tabular}{lccccccc}
        \toprule
        \multirow{2}{*}{\textbf{Agent Model}} & \multicolumn{2}{c}{\textbf{Mobile}} & \multicolumn{2}{c}{\textbf{Desktop}} & \multicolumn{2}{c}{\textbf{Web}} & \multirow{2}{*}{\cellcolor{white}\textbf{Avg}} \\
        \cmidrule(lr){2-3} \cmidrule(lr){4-6} \cmidrule(lr){6-7} 
         & \textbf{Text} & \textbf{Icon/Widget} & \textbf{Text} & \textbf{Icon/Widget} & \textbf{Text} & \textbf{Icon/Widget} & \cellcolor{white}{} \\
        \midrule
        SeeClick \cite{cheng2024seeclick} & 78.4 & 50.7 & 70.1 & 29.3 & 55.2 & 32.5 & \textbf{55.1} \\
        OmniParser-v2 \cite{yu2025omniparser} & 95.5 & 74.6 & 92.3 & 60.9 & 88.0 & 59.6 & \textbf{80.7} \\
        Qwen2.5-VL-3B \cite{bai2025qwen2} & 93.4 & 73.5 & 88.1 & 58.6 & 88.0 & 71.4 & \textbf{80.9} \\
        UI-TARS-2B \cite{qin2025ui} & 95.2 & 79.1 & 90.7 &  68.6 &  87.2 & 78.3 & \textbf{84.7} \\
        OS-Atlas-Base-7B \cite{wu2024atlas} & 96.2 & 83.4 & 89.7 & 69.3 & 94.0 & 79.8 & \textbf{87.1} \\
        Jedi-3B \cite{xie2025scaling} & 96.6 & 81.5 & 96.9 & 78.6 & 88.5 & 83.7 & \textbf{88.6} \\
        Qwen2.5-VL-7B \cite{bai2025qwen2} & 97.6 & 87.2 & 90.2 & 74.2 & 93.2 & 81.3 & \textbf{88.8} \\
        UI-TARS-1.5-7B \cite{qin2025ui} & 95.9 & 84.8 & 94.9 & 80.7 & 90.6 & 86.2 & \textbf{89.7} \\
        UI-TARS-72B \cite{qin2025ui} & 94.8 & 86.3 & 91.2 & 87.9 & 91.5 & 87.7 & \textbf{90.3} \\
        UI-TARS-7B \cite{qin2025ui} & 96.9 & 89.1 & 95.4 & 85.0 & 93.6 & 85.2 & \textbf{91.6} \\
        Jedi-7B \cite{xie2025scaling} & 96.9 & 87.2 & 95.9 & 87.9 & 94.4 & 84.2 & \textbf{91.7} \\
        GUI-Actor-7B \cite{wu2025gui}& 97.6 & 88.2 & 96.9 & 85.7 & 93.2 & 86.7 & \textbf{92.1} \\
        \midrule
        \rowcolor{blue!15} Qwen2.5-VL-7B \cite{bai2025qwen2} & 98.3 & 86.7 &  89.4 & 63.3 & 92.7 & 81.8 & \textbf{88.8} \\
        \rowcolor{red!15} + AutoFocus &  98.8 & 88.9  & 95.6  &  85.5 &  96.2 & 91.2  &  \textbf{93.7}  \\
        \rowcolor{blue!15} Qwen2.5-VL-72B \cite{bai2025qwen2} & 99.0 & 90.1 & 96.4 & 87.1 & 96.6 & 90.6 & \textbf{94.0} \\
        \rowcolor{red!15}+ AutoFocus  & 99.0  &  93.2  & 97.5  & 89.3  & 96.4  &  92.7  &  \textbf{95.5} \\
        \rowcolor{blue!15} UI-Venus-7B \cite{gu2025ui} & 99.0 & 90.0 & 97.0& 90.7& 96.2 &88.7 &\textbf{94.1}\\
        \rowcolor{red!15}+ AutoFocus  & 99.3  &  91.4  & 97.5  & 90.7  & 96.8  &  91.2  &  \textbf{95.3} \\
        \bottomrule
    \end{tabular}}
    \label{tab:screenspot_v2_comparison}
\end{table}

\subsection{Ablation Studies}

To rigorously understand which design choices contribute to AutoFocus, we conduct ablations along two axes: 
(i) \textbf{component-level} ablations to validate the necessity of each module in the pipeline, and 
(ii) \textbf{hyperparameter} ablations to quantify sensitivity to inference-time settings, \eg, sampling budget and zoom scale. 
Unless otherwise specified, we report \emph{Accuracy} on ScreenSpot-Pro and ScreenSpot-V2 to ensure fair compute comparisons.

\subsubsection{Component ablations (What matters?).}
To dissect the contribution of each module within the AutoFocus framework, we evaluate a series of ablation variants by progressively integrating or isolating its key components. This systematic analysis, summarized in~\cref{tab:ablation_results}, reveals a clear evolutionary trajectory from deterministic point prediction to uncertainty-aware region refinement.
\begin{wraptable}{t}{0.5\textwidth}
    \centering
    \small
    \vspace{-18pt} 
    \setlength{\tabcolsep}{4pt} 
    \renewcommand{\arraystretch}{1.1}
    \caption{\textbf{Component ablation study of AutoFocus.} Accuracy (\%) on ScreenSpot-Pro/V2 with Qwen2.5-VL-7B.}
    \label{tab:ablation_results}
    \begin{tabular}{lcc}
        \toprule
        \textbf{Variant} & \textbf{SS-Pro} $\uparrow$ & \textbf{SS-V2} $\uparrow$ \\
        \midrule
        Baseline (no zoom) & 26.8 & 88.8 \\
        + Multi-sample only & 27.5 & 89.4 \\
        + Global proposals & 31.6 & 90.9 \\
        + Local proposals & 33.1 & 92.3 \\
        \midrule
        w/o Axial PPL & 34.1 & 92.9 \\
        w/o SAZ & 33.8 & 92.4 \\
        w/o Vis-Aggregation & 31.6 & 91.1 \\
        \midrule
        \textbf{AutoFocus (Full)} & \textbf{35.4} & \textbf{93.7} \\
        \bottomrule
    \end{tabular}\vspace{-10pt}
\end{wraptable}

We first establish a \textbf{Baseline (no zoom)} using standard single-pass coordinate prediction, which achieves a modest 26.8\% accuracy on ScreenSpot-Pro, severely constrained by the resolution bottleneck. Simply introducing \textbf{+ Multi-sample only}, where we sample $N$ candidates but execute the most confident one without any spatial modeling, yields marginal gains (+0.7\%), suggesting that stochastic sampling alone cannot resolve grounding ambiguities. 
A significant performance leap occurs with the introduction of the probability density field and zoom-in strategy. Transitioning to the \textbf{+ Global proposals} provides a +4.1\% boost over the sampling baseline, while \textbf{+ Local proposals} improves accuracy to 33.1\%. These results validate the effectiveness of our zoom-in strategy, demonstrating that shifting the model's focus from discrete points to structured regional hypotheses is fundamental to overcoming high-resolution visual clutter.

We further ablate the sophisticated design choices that distinguish AutoFocus from naive zooming mechanisms:
\begin{itemize}
\item \textbf{Importance of Structural Priors (w/o Axial PPL):} Replacing directional uncertainty $(\mathcal{L}_{ppl}^x, \mathcal{L}_{ppl}^y)$ with a global sequence perplexity leads to a performance drop (34.1\% vs. 35.4\%). This validates that GUI elements possess strong structural anisotropy; modeling $x$ and $y$ uncertainty independently is crucial for effectively capturing elongated elements such as text bars or vertical menus.
\item \textbf{Precision vs. Context (w/o SAZ):} Disabling Shape-Aware Zooming forces the model to use raw, uncertainty-derived crops. The resulting degradation (33.8\%) highlights the necessity of SAZ in balancing character-level precision with the semantic context required for high-level reasoning.
\item \textbf{Rational Decision Making (w/o Visual Aggregation):} The most dramatic drop occurs when replacing our multi-region visual aggregation with naive confidence-based selection, with accuracy falling back to 31.6\%. 

This suggests that the ``second look'' provided by AutoFocus is effective when regional insights are synthesized rather than merely filtered, proving that visual aggregation is a cornerstone of our refinement logic.
\end{itemize}

\subsubsection{Hyperparameter ablations (How sensitive?).}
\paragraph{Sampling budget $N$.}
\begin{wrapfigure}{r}{0.5\textwidth}
  \centering
  \includegraphics[width=\linewidth]{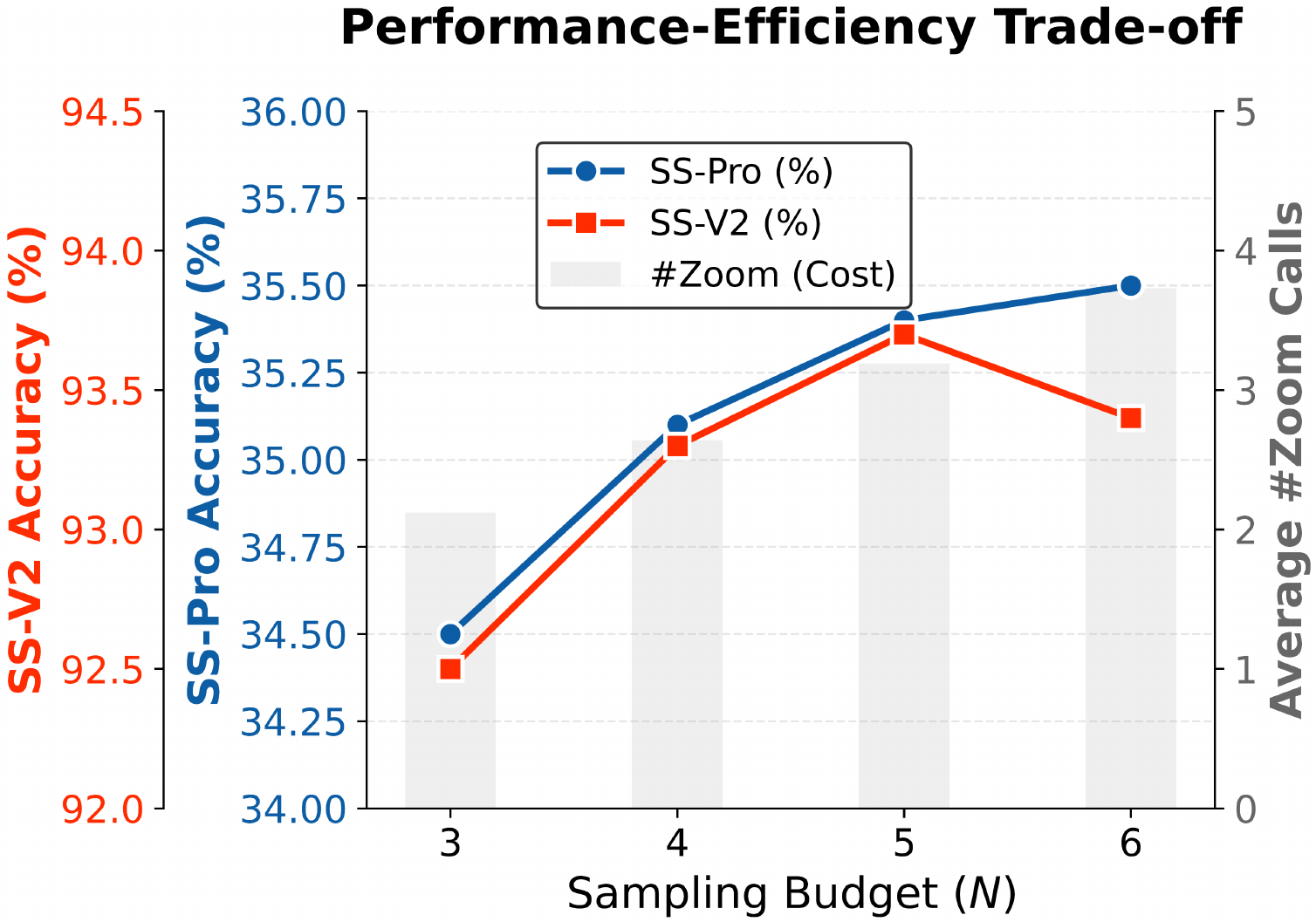}
  \caption{Effect of Sampling Budget $N$.}
  \label{fig:comparison}
  \vspace{-20pt} 
\end{wrapfigure}
\cref{fig:comparison} analyzes the effect of the stochastic sampling budget $N$ on performance and computational cost (\eg, average zoom calls).
As $N$ increases from 3 to 5, accuracy on ScreenSpot-Pro steadily improves (34.5\% $\rightarrow$ 35.4\%), while ScreenSpot-V2 shows similar gains (92.5\% $\rightarrow$ 93.7\%). This confirms that multiple hypothesis sampling enhances spatial uncertainty estimation and proposal diversity.

However, increasing $N$ beyond 5 yields only marginal improvements (+0.1\% on SS-Pro) while noticeably increasing the average number of zoom operations. This indicates diminishing returns when excessive hypotheses introduce redundant spatial regions. 
We therefore adopt $N=5$ as a balanced trade-off between accuracy and inference efficiency.

\begin{wraptable}{r}{0.5\textwidth} 
  \centering
  \vspace{-30pt} 
  \caption{Effect of squareness factor $\lambda$.}
  \label{tab:sub_scale}
  \small 
  \setlength{\tabcolsep}{6pt}
  \begin{tabular}{lcc}
    \toprule
    \textbf{$\lambda$} & \textbf{SS-Pro} $\uparrow$ & \textbf{SS-V2} $\uparrow$ \\
    \midrule
    0      & 33.8 & 92.4 \\
    0.25   & 34.4 & 93.0 \\
    \textbf{0.5}    & \textbf{35.4} & \textbf{93.7} \\
    0.75   & 35.3 & 93.5 \\
    \bottomrule
  \end{tabular}
  \vspace{-10pt} 
\end{wraptable}
\paragraph{Squareness factor $\lambda$.}
~\cref{tab:sub_scale} evaluates the sensitivity to the Shape-Aware Zooming parameter $\lambda$, which interpolates between anisotropic crops ($\lambda=0$) and square crops ($\lambda=1$).
When $\lambda=0$, the model relies purely on uncertainty-derived anisotropic regions, resulting in suboptimal accuracy (33.8\% on SS-Pro), likely due to insufficient contextual coverage.
Moderate values improve performance, with $\lambda=0.5$ achieving the best overall results on both benchmarks.
Larger values (\eg, $\lambda=0.75$) slightly degrade performance, suggesting that overly square crops introduce unnecessary background regions and dilute high-resolution detail.
These results validate the importance of balancing structural precision and contextual completeness, and demonstrate that AutoFocus is relatively robust within a reasonable $\lambda$ range.
\section{Conclusion}
We revisit high-resolution GUI grounding through the lens of uncertainty. Instead of introducing additional supervision or complex training strategies, AutoFocus leverages token-level perplexity as an intrinsic signal to guide adaptive visual refinement. By converting generative uncertainty into structured spatial modeling, our approach enables principled test-time search without additional training. The consistent gains across models and benchmarks indicate that internal generative uncertainty can serve as a practical foundation for scalable visual reasoning beyond single-pass inference.

%
%
\bibliographystyle{splncs04}
\bibliography{main}
\end{document}